\documentclass[11pt,a4paper]{article}
\usepackage[hyperref]{emnlp2020}
\usepackage{times}
\usepackage{latexsym}

\usepackage{booktabs}
\usepackage{subcaption}
\usepackage{graphicx} 
\usepackage{latexsym}
\usepackage{amsmath}
\usepackage{bbm}
\usepackage{lipsum}\usepackage{tikz-dependency}
\usepackage{enumitem}
\usepackage{footnote}
\usepackage{tabularx}
\usepackage{multirow}
\usepackage{stfloats}
\usepackage{todonotes}
\usepackage{CJKutf8}
\usepackage[utf8]{vietnam}
\makesavenoteenv{tabular}

\newcommand{\Sref}[1]{\S\ref{#1}}

\usepackage{microtype}

\aclfinalcopy 
\def\aclpaperid{9} 

\title{{U}niversal {D}ependencies {A}ccording to {BERT}: {B}oth {M}ore {S}pecific and {M}ore {G}eneral}

\author{Tomasz Limisiewicz \and Rudolf Rosa \and David Mare\v{cek}\\
    Institute of Formal and Applied Linguistics, Faculty of Mathematics and Physics \\
    Charles University, Prague, Czech Republic \\
  \texttt{\{limisiewicz,rosa,marecek\}@ufal.mff.cuni.cz}
}
\date{}

\begin{document}
\maketitle
\begin{abstract}
This work focuses on analyzing the form and extent of syntactic abstraction captured by BERT by extracting labeled dependency trees from self-attentions.

Previous work showed that individual BERT heads tend to encode particular dependency relation types. We extend these findings by explicitly comparing BERT relations to Universal Dependencies (UD) annotations, showing that they often do not match one-to-one.
We suggest a method for relation identification and syntactic tree construction. Our approach produces significantly more consistent dependency trees than previous work, showing that it better explains the syntactic abstractions in BERT.

At the same time, it can be successfully applied with only a minimal amount of supervision and generalizes well across languages.
\end{abstract}

\section{Introduction and Related Work}

In recent years, systems based on Transformer architecture achieved state-of-the-art results in language modeling \cite{devlin2018bert} and machine translation \cite{vaswani2017attention}. Additionally, the contextual embeddings obtained from the intermediate representation of the model brought improvements in various NLP tasks. Multiple recent works try to analyze such latent representations \citep{blackboxnlp-2019}, observe
syntactic properties in some Transformer self-attention heads,
and extract syntactic trees from the attentions matrices
\citep{raganato-tiedemann-2018-analysis,marecek-rosa-2019-balustrades,clark2019does, jawahar-etal-2019-bert}.

In our work, we focus on the comparative analysis of the syntactic structure,
examining how the BERT self-attention weights correspond to Universal Dependencies (UD) syntax \citep{nivre2016universal}.
We confirm the findings of \citet{vig-belinkov-2019-analyzing} and \citet{voita-etal-2019-analyzing}
that in Transformer based systems particular heads tend to capture specific dependency relation types
(e.g. in one head the attention at the predicate is usually focused on the nominal subject).

We extend understanding of syntax in BERT by examining the ways in which it systematically diverges from
standard annotation (UD).
We attempt to bridge the gap between them in three ways:

\begin{itemize}
    \item We modify the UD annotation of three linguistic phenomena
    to better match the BERT syntax (\Sref{sec:udadaptation})

    \item We introduce a head ensemble method, combining multiple heads which capture the same dependency relation label (\Sref{sec:headensemble})

    \item We observe and analyze multipurpose heads, containing multiple syntactic functions (\Sref{sec:multipurpose})

\end{itemize}

Finally, we apply our observations to improve the method of extracting dependency trees from attention (\Sref{sec:parsing}),
and analyze the results both in a monolingual and a multilingual setting (\Sref{sec:results}).

Our method crucially differs from probing \cite{belinkov-etal-2017-evaluating,hewitt2019structural,chi2020finding, kulmizev-etal-2020-neural}. We do not use treebank data to train a parser;
rather, we extract dependency relations directly from selected attention heads.
We only employ syntactically annotated data to select the heads; however, this means estimating relatively few parameters, and only a small amount of data is sufficient for that purpose
(\Sref{sec:supervision}).

\section{Models and Data}

We analyze the uncased base BERT model for English, which we will refer to as {\bf enBERT},
and the uncased multilingual BERT model, {\bf mBERT}, for English, German, French, Czech, Finnish, Indonesian, Turkish, Korean, and Japanese \footnote{Pretrained models are available at \url{https://github.com/google-research/bert}}. The code shared by \citet{clark2019does} \footnote{\url{https://github.com/clarkkev/attention-analysis}} substantially helped us in extracting attention weights from BERT.

To find syntactic heads, we use: 1000 EuroParl multi parallel sentences~\cite{koehn-2004} for five European languages, automatically annotated with UDPipe UD 2.0 models \cite{udpipe:2017};
Google Universal Dependency Treebanks (GSD) for Indonesian, Korean, and Japanese \cite{mcdonald-etal-2013-universal}; the UD Turkish Treebank (IMST-UD) \cite{sulubacak-etal-2016}.

We use another PUD treebanks from the CoNLL 2017 Shared Task for evaluation of mBERT in all languages \cite{11234/1-2184}\footnote{Mentioned treebanks are available at the Universal Dependencies web page \url{https://universaldependencies.org}}.


\section{Adapting UD to BERT}
\label{sec:udadaptation}

Since the explicit dependency structure is not used in BERT training, syntactic dependencies captured in latent layers are expected to diverge from annotation guidelines. After initial experiments, we have observed that some of the differences are systematic (see Table~\ref{tab:modifications}). 

\begin{savenotes}
\newcolumntype{C}{ >{\arraybackslash} m{1.3cm} }
\newcolumntype{T}{ >{\centering\arraybackslash} m{4cm} }
\begin{table}[h]
\small
\begin{tabular}{|C|C|T|}
\toprule
UD & Modified &  Example  \\ \midrule
Copula attaches to a noun & Copula is a root.
\footnote{Certain dependents of the original root (e.g., subject, auxiliaries) are rehanged and attached to the new root -- copula verb.} 
& 
\begin{dependency}[edge unit distance=1.5ex]
    \begin{deptext}[column sep=0.5cm]
     cat\& {\bf is}\& an\& animal\\
\end{deptext}
\deproot{1}{root}
\depedge{1}{2}{cop}
\depedge{1}{4}{nsubj}
\depedge[edge style={blue!}, edge below]{2}{1}{nsubj}
\deproot[edge style={blue!}, edge below]{2}{root}
\depedge[edge style={blue!}, edge below]{2}{4}{obj}
\end{dependency} \\ \midrule
 Expletive is not a subject & Expletive is treated as a subject  &
\begin{dependency}[edge unit distance=1.5ex]
    \begin{deptext}[column sep=0.5cm]
     {\bf there}\& is\& a\& spoon\\
\end{deptext}
\depedge{2}{1}{expl}
\depedge{2}{4}{nsubj}
\depedge[edge style={blue!}, edge below]{2}{1}{nsubj}
\depedge[edge style={blue!}, edge below]{2}{4}{obj}
\end{dependency}  \\ \midrule
In multiple coordination, all conjuncts attach to the first conjunct & Conjunct attaches to a previous one & 
\begin{dependency}[edge unit distance=1.ex]
    \begin{deptext}[column sep=0.07cm]
     apples\& ,\& oranges\& and\& {\bf pears}\\
\end{deptext}
\depedge{1}{3}{conj}
\depedge{1}{5}{conj}
\depedge[edge style={blue!}, edge below]{1}{3}{conj}
\depedge[edge style={blue!}, edge below]{3}{5}{conj}
\end{dependency} \\ \bottomrule
\end{tabular}
    \caption{Comparison of original Universal Dependencies annotations (\textbf{edges above}) and our modification (\textcolor{blue}{edges below}).}
    \label{tab:modifications}
\end{table}
\end{savenotes}

Based on these observations, we modify the UD annotations in our experiments
to better fit the BERT syntax, using UDApi\footnote{\url{https://udapi.github.io}} \cite{popel-etal-2017-udapi}.

The main motivation of our approach is to get trees similar to structures emerging from BERT, which we have observed in qualitative analysis of attention weights. We note that for copulas and coordinations, BERT syntax resembles Surface-syntactic UD (SUD) \citep{gerdes:hal-01930614}. Nevertheless, we decided to use our custom modification, since some systematic divergences between SUD and the latent representation occur as well. It is not our intention to compare two annotation guidelines.
A comprehensive comparison between extracting UD and extracting SUD trees from BERT was performed by \cite{kulmizev-etal-2020-neural}. However, they used a probing approach, which is noticeably different from our setting.


\section{Head Ensemble}
\label{sec:headensemble}

\begin{figure*}
    \centering
    \includegraphics[width=15cm]{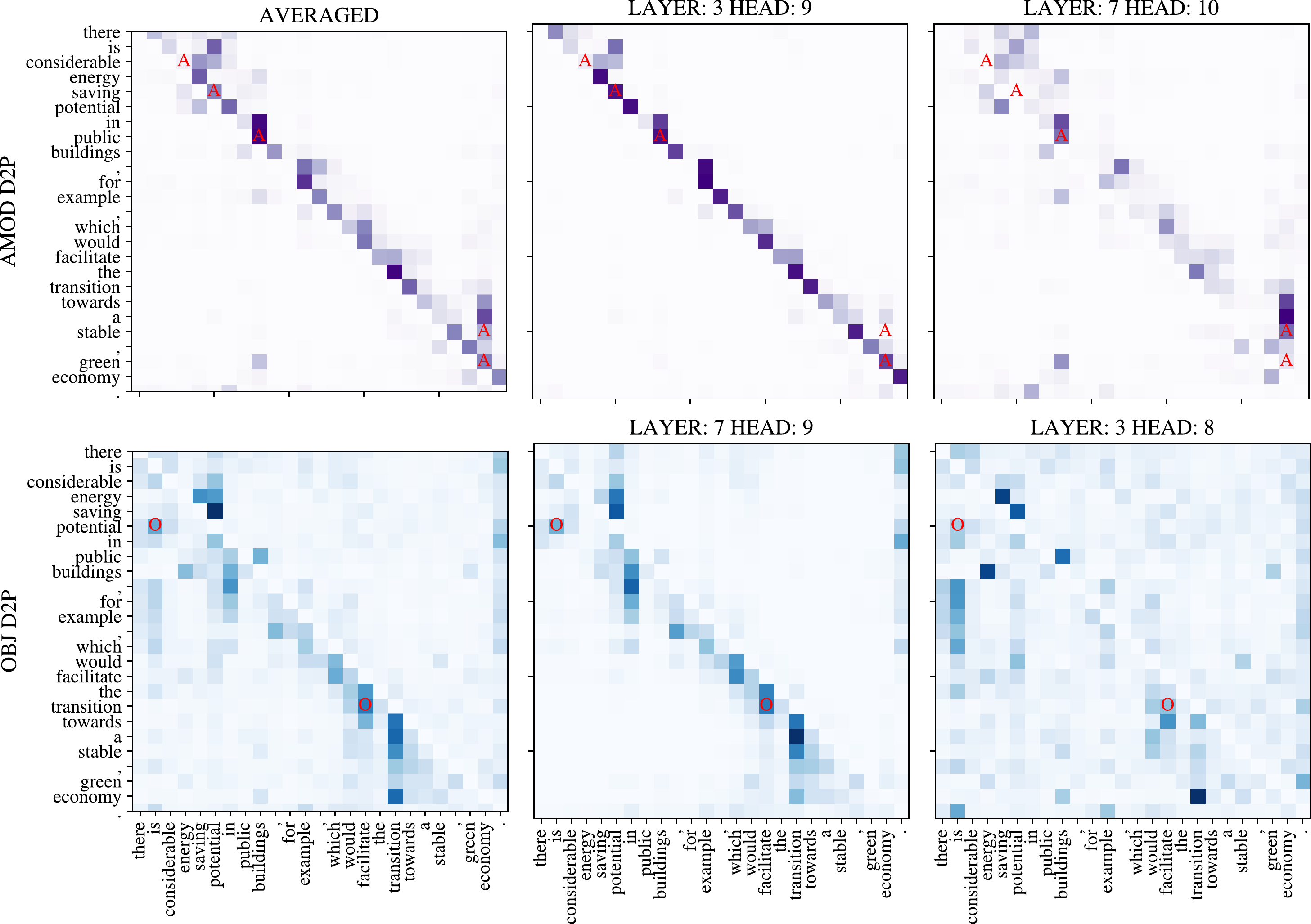}
    \caption{Examples of two enBERT's attention heads covering the same relation label and their average. Gold relations are marked by red letters.
    } \label{fig:attentions}
\end{figure*}
In line with \citet{clark2019does} and other studies \citet{voita-etal-2019-analyzing, vig-belinkov-2019-analyzing}, we have noticed that a specific syntactic relation type can often be found in a specific head.
Additionally, we observe that a single head often captures only a specific aspect or subtype of one UD relation type,
motivating us to combine multiple heads to cover the full relation.

Figure~\ref{fig:attentions} shows attention weights of two syntactic heads (right columns) and their average (left column).
In the top row (purple), both heads identify the parent noun for an adjectival modifier:
Head 9 in Layer 3 if their distance is two positions or less, Head 10 in Layer 7 if they are further away (as in ``a \underline{stable} , green \underline{economy}'').

Similarly, for an object to predicate relation (blue bottom row), Head 9 in Layer 7 and Head 8 in Layer 3 capture pairs with shorter and longer positional distances, respectively.

\subsection{Dependency Accuracy of Heads}
\label{sec:accuracy}

To quantify the amount of syntactic information conveyed by a self-attention head $A$ for a dependency relation label $l$ in a specific direction $d$ (for instance predicate $\rightarrow$ subject), we compute:

\begin{equation*}
DepAcc_{l,d, A} =\frac{|\{(i,j) \in E_{l,d} : j = \arg\max A[i] \}|}{|E_{l,d}|},
\end{equation*}
where $E_{l,d}$ is a set of all dependency tree edges with the label $l$ and with direction $d$, i.e., in dependent to parent direction (abbreviated to \textbf{p2d}) the first element of the tuple $i$ is dependent of the relation and the second element $j$ is the governor; $A[i]$ is the $i^{th}$ row of the attention matrix $A$.


In this article, when we say that head with attention matrix $A$ is syntactic for a relation type $l$, we mean that its $DepAcc_{l,d,A}$ is high in one of the directions (parent to dependent \textbf{p2d} or dependent to parent \textbf{d2p}).

\subsection{Method}

Having observed that some heads convey only partial information about a UD relation, we propose a method to connect knowledge of multiple heads. 

Our objective is to find a set of heads for each directed relation so that their attention weights after averaging
have a high dependency accuracy. The algorithm is straightforward:  we define the maximum number $N$ of heads in the subset; sort the heads based on their $DepAcc$ on development set; starting from the most syntactic one we check whether including head's attention matrix in the average would increase $DepAcc$; if it does the head is added to the ensemble. When there are already $N$ heads in the ensemble, the newly added head may substitute another added before, so to maximize $DepAcc$ of the averaged attention matrices.\footnote{The code is available at GitHub: \url{https://github.com/Tom556/BERTHeadEnsembles}}

We set $N$ to be 4, as allowing larger ensembles does not improve the results significantly. 



\section{Dependency Tree Construction}
\label{sec:parsing}
To extract dependency trees from self-attention weights, we use a method similar to \citet{raganato-tiedemann-2018-analysis},
which employs a maximum spanning tree algorithm \citep{edmonds-branching} and uses gold information about the root of the syntax tree.


We use the following steps to construct a labeled dependency tree:
\begin{enumerate}  
\item For each non-clausal UD relation label, syntactic heads ensembles are selected as described in Section~\ref{sec:headensemble}. Attention matrices in the ensembles are averaged. Hence, we obtain two matrices for each label (one for each direction: "dependent to parent" and "parent to dependent")
\item The "dependent to parent" matrix is transposed and averaged with "parent to dependent" matrix. We use a weighted geometric average with weights corresponding to dependency accuracy values for each direction. \label{direction-averaging}
\item We compute the final dependency matrix by max-pooling over all individual relation-label matrices from step \ref{direction-averaging}.
At the same time, we save the syntactic-relation label that was used for each position in the final matrix.\label{max-pooling}
\item In the final matrix, we set the row corresponding to the gold root to zero, to assure it will be the root in the final tree as well.
\item We use the Chu-Liu-Edmond's algorithm~\citep{edmonds-branching} to find the maximum spanning tree. For each edge, we assign the label saved in step 3.
\end{enumerate}

It is important to note that the total number of heads used for tree construction can be at most
$4 * 12 * 2 = 96$, (number of heads per ensemble $*$ number of considered labels $*$ two directions). 
However, the number of used heads is typically much lower (see Table~\ref{tab:uas}). 
That means our method uses at most 96 integer parameters (indices of the selected heads), considerably less than projection layers in fine-tuning or structural probing, consisting of thousands of real parameters.

As far as we know, we are first to construct labeled dependency trees from attention matrices in Transformer. Moreover, we have extended the previous approach by using an ensemble of heads instead of a single head.

\begin{savenotes}
\begin{flushleft}
\begin{table}
\small
\begin{tabular}{@{}llllll@{}}
\toprule
Relation & Base- & \multicolumn{2}{c}{1 Head} & \multicolumn{2}{c}{4 Heads}   \\ 
label & line & d2p & p2d & d2p & p2d \\ \midrule
amod & 78.3 & 90.6 & 77.5 & \bf93.8 & 79.5 \\
advmod & 48.7 & 53.3 & 62.0 & 62.1 & \bf63.6 \\
aux & 69.2 & 90.9 & 86.9 & \bf94.5 & 88.0 \\
case & 36.4 & 83.0 & 67.1 & \bf88.4 & 68.9 \\
compound & 75.8 & 83.2 & 75.8 & \bf87.0 & 79.1 \\
conjunct & 31.7 & 47.4 & 41.6 & \bf58.8 & 51.3 \\
det & 56.5 & 95.2 & 62.3 & \bf97.2 & 69.4 \\
nmod & 25.4 & 34.3 & 41.5 & 49.1 & \bf54.7 \\
nummod & 57.9 & 75.9 & 64.6 & \bf79.3 & 72.6 \\
mark & 53.7 & 66.2 & 54.7 & \bf73.5 & 65.9 \\
obj\footnote{Objects also include indirect objects (\textit{iobj}).} & 39.2 & 84.9 & 68.6 & \bf89.3 & 78.5 \\
nsubj & 45.8 & 56.2 & 62.7 & 57.8 & \bf76.0 \\ \midrule
$\Uparrow$ AVG. \\ NON-CLAUSAL  & 52.8 & \multicolumn{2}{c}{67.8} & \multicolumn{2}{c}{\bf74.1} \\ \midrule
acl & 27.9 & 41.5 & 36.5 & \bf50.5 & 43.8 \\
advcl & 9.3 & 26.3 & 26.7 & \bf40.7 & 26.3 \\
csubj & 20.0 & 20.7 & \bf31.0 & 24.1 & \bf31.0 \\
x/ccomp\footnote{Open clausal complements and clausal complements.} & 34.8 & 60.4 & 47.9 & \bf66.9 & 52.1 \\
parataxis & 10.4 & 17.6 & 12.1 & 23.1 & \bf24.2 \\ \midrule
$\Uparrow$ AVG. CLAUSAL & 20.5 & \multicolumn{2}{c}{32.1} & \multicolumn{2}{c}{\bf38.3} \\ \midrule
punct & 9.4 & 21.1 & 40.3 & 28.4 & \bf44.0 \\
dep\footnote{\textit{Dep} relations and all relations not included in this table.} & 18.8 & 21.6 & 33.1 & 25.1 & \bf37.0 \\

 \bottomrule
\end{tabular}
\caption{Dependency accuracy for single heads, 4 heads ensembles, and positional baselines. The evaluation was done using the pretrained model enBERT and modified UD as described in Section~\ref{sec:udadaptation}.} 
\label{tab:depacc}
\end{table}
\end{flushleft}
\end{savenotes}

\section{Results}
\label{sec:results}

\subsection{Dependency Accuracy}
\label{sec:depacc-results}
In Table~\ref{tab:depacc}, we present results for the dependency accuracy (Section~\ref{sec:accuracy}) of a single head, four heads ensemble, and the positional baseline.\footnote{The positional baseline looks at the most frequent relative position for each dependency label \citep{voita-etal-2019-analyzing}.}

Noticeably, a single attention head surpasses the baseline for every relation label in at least one direction. The average of 4 heads surpasses the baseline by more than 10\% for every relation.

Ensembling brings the most considerable improvement for nominal subjects (p2d: +13.3 pp) and noun modifiers (p2d: +13.2 pp). The relative change of accuracy is more evident for clausal relations than non-clausal. Dependent to parent direction has higher accuracy for modifiers (except adverbial modifiers), functional relations, and objects, whereas parent to dependent favors other nominal relations (nominal subject and nominal modifiers).

Introducing the UD modifications (Section~\ref{sec:udadaptation}) had a significant effect for nominal subject.
Without such modifications, the accuracy for parent to dependent direction would drop from 76.0\% to 70.1\%




\begin{figure}[h]
    \begin{subfigure}{\linewidth}
        \centering
        \includegraphics[width=\linewidth]{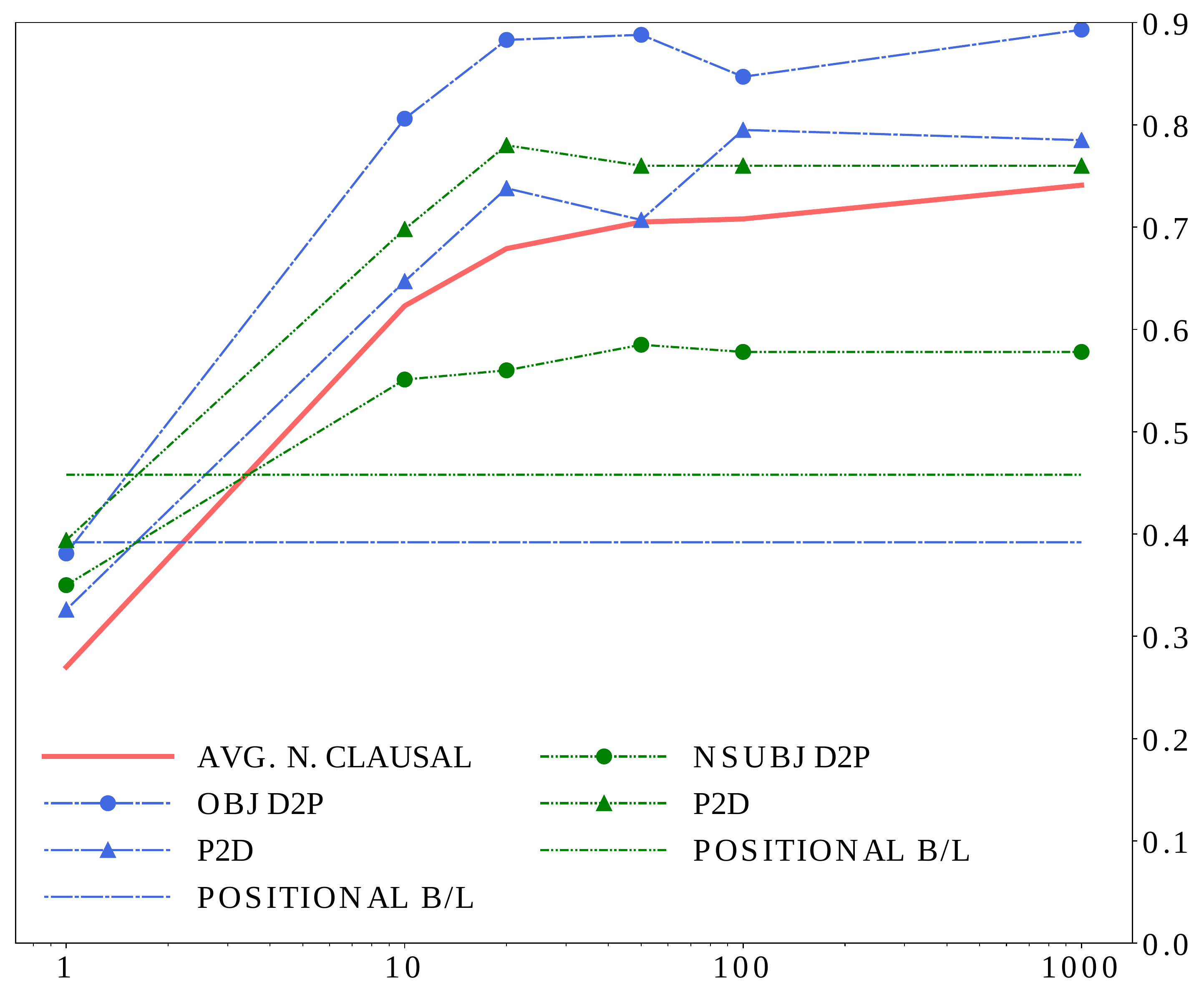}
        \caption{Non-clausal relations} \label{fig:daNom}
    \end{subfigure}
    \begin{subfigure}{\linewidth}
        \centering
        \includegraphics[width=\linewidth]{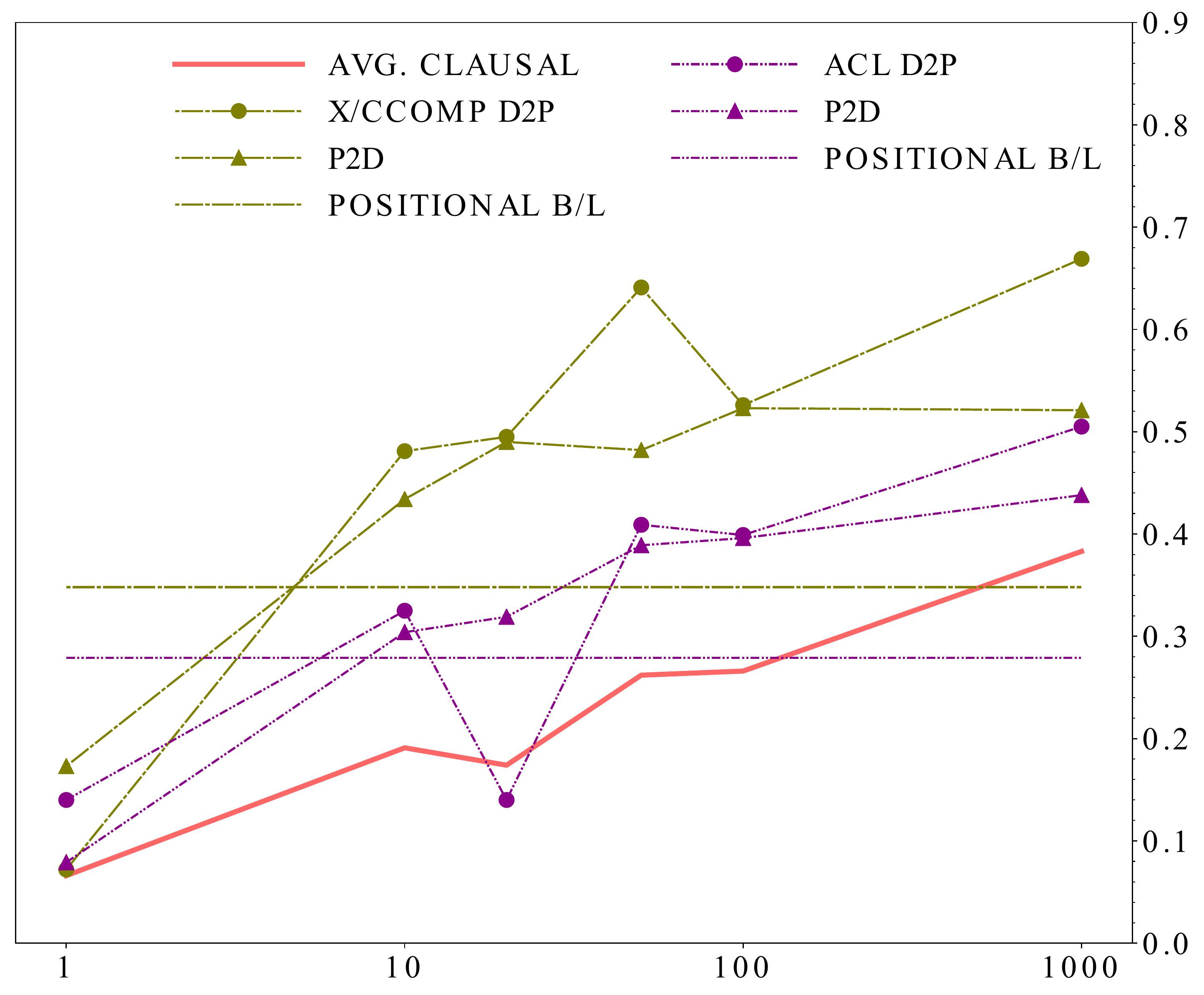}
        \caption{Clausal relations} \label{fig:daClause}
    \end{subfigure}
    \caption{Dependency accuracy against the number of sentences used for selection.}
    \label{fig:daSize}
\end{figure}

\paragraph{Selection Supervision}
\phantomsection
\label{sec:supervision}
The selection of syntactic heads requires annotated data for accuracy evaluation.
In Figure \ref{fig:daSize}, we examine what number of annotated sentences is sufficient,
using 1, 10, 20, 50, 100 or 1000 sentences.

For non-clausal relations (Figure~\ref{fig:daNom}),
head selection on just 10 annotated sentences allows us to surpass the positional baseline. Using over 20 examples brings only a minor improvement. For clausal relations (Figure~\ref{fig:daClause}), the score improves steadily with more data. However, even for the full corpus, it is relatively low, since the clausal relations are less frequent in the corpus and harder to identify due to longer distances between dependent and parent.

\begin{savenotes}
\begin{table*}[]
\centering
\begin{tabular}{@{}lccccccc@{}}
\toprule
Setting & Use labels & Model  & Selection & Heads per & Heads &  UAS & LAS \\ 
 &  &  & sentences & ensemble & used & & \\\midrule
Left branching baseline & --- & --- & --- & --- &  --- & 11.0 & --- \\
Right branching baseline & --- & --- & --- & --- & ---& 35.5 & --- \\\midrule
Raganato+ (paper) & no  & NMT    & 1000* & --- & 1  & 38.9 & --- \\
Raganato+           & no  & enBERT & 1000*  & ---  & 1 & 37.2 & --- \\\midrule
\multirow{4}{*}{Our method} & no  & enBERT & 1000 & 1 & 2 & 36.0 & --- \\
         & yes & enBERT & 1000 &  1 & 15  & 37.4 & 9.5 \\
         & yes & enBERT &  20 & 4 & 36 & 43.6 & 14.5 \\
         & no  & enBERT &  1000 & 4 & 8 & 51.2 & --- \\ \midrule
Our method & yes & enBERT & 1000 &  4 & 48 &  \bf52.0 & \bf21.7 \\ \bottomrule
\end{tabular}
    \caption{Evaluation results for different settings of dependency trees extraction. 
    UD modifications were not applied here. (*In Raganato+ experimens, the trees were induced from each encoder head, but we report only the results for the head with the highest UAS on 1000 test sentences.)}
    \label{tab:uas}
\end{table*}
\end{savenotes}

\subsection{Dependency Tree Construction}
\phantomsection
In Table~\ref{tab:uas}, we report the evaluation results on the English PUD treebank~\citep{11234/1-2184} using unlabeled and labeled attachment scores (UAS and LAS). For comparison, we also include the left- and right-branching baseline with gold root information, and the highest score obtained by \citet{raganato-tiedemann-2018-analysis} who used the neural machine translation Transformer model and extracted whole trees 
from a single attention head.
Also, they did not perform direction averaging. The results show that ensembling multiple attention heads for each relation label allows us to construct much better trees than the single-head approach.\footnote{To assure comparability we do not modify the UD annotation for the results in this table.}

The number of unique heads used in the process turned out to be two times lower than the maximal possible number (96). This is because many heads appear in multiple ensembles. We examine it further in Section~\ref{sec:multipurpose}.

Furthermore, to the best of our knowledge, we are the first to produce labeled trees and report both UAS and LAS.

Just for reference, the recent unsupervised parser~\cite{han-etal-2019-enhancing} obtains 61.4\% UAS. However, the results are not comparable since the parser uses information about gold POS tags, and the results were measured on different evaluation data (WSJ Treebank).

\paragraph{Ablation}

We analyze how much the particular steps described in Section~\ref{sec:parsing} influenced the quality of constructed trees. We also repeat the experimental setting proposed by \citet{raganato-tiedemann-2018-analysis} on enBERT model to see whether a language model is better suited to capture syntax than a translation system. Additionally, we alter the procedure described in Section~\ref{sec:parsing} to analyze which decision influenced our results the most, i.e., we change:
\begin{itemize}
\item Size of head ensembles 
\item Number of sentences used for head selection
\item Use the same head ensemble for all relation labels in each direction. Hence we do not conduct max-pooling described in section~\ref{sec:parsing}, point~\ref{max-pooling}.
\end{itemize}

In Table~\ref{tab:uas}, we see that the method by~\citet{raganato-tiedemann-2018-analysis} applied to enBERT produces slightly worse trees than the same method applied to neural machine translation.
If we do not use ensembles and only one head per each relation label and direction is used, our pipeline from Section~\ref{sec:parsing}
offers only 0.2 pp rise in UAS and poor LAS.
The analysis shows that the introduction of head ensembles of size four has brought the most significant improvement in our method of tree construction, which is roughly +15 pp for both the variants (with and without labels).

Together with the findings in Section~\ref{sec:depacc-results} this supports our claim that syntactic information is spread across many Transformer's heads. Interestingly, max-pooling over labeled matrices improve UAS only by 0.8 pp. Nevertheless, this step is necessary to construct labeled trees. The performance is competitive, even with as little as 20 sentences used for head selection, which is in line with our findings from Section~\ref{sec:supervision}.

\begin{savenotes}
\begin{table}[t]
\small
\centering
\begin{tabular}{@{}ccccccc@{}}
\toprule
Lang- &  Features & \multicolumn{2}{c}{DepAcc} & \multicolumn{2}{c}{UAS} & LAS   \\ 
uage &   & b-line & Our & b-line & Our & Our \\ \midrule
EN & SVO, AN & 52.8 & \bf73.2 & 35.5 & \bf51.0  & 21.8 \\
DE & ---\footnote{No dominant order}, AN & 42.3 & \bf72.9  & 32.9 & \bf45.5 & 19.5 \\
FR  &  SVO, NA & 50.6 & \bf72.8  & 34.7 & \bf48.3 & 18.0\\
CS &  SVO, AN &44.3 & \bf69.7  & 34.0 & \bf40.1 &17.1 \\
FI & SVO, AN&  55.6 & \bf77.0  & 35.5 & \bf45.8 & 15.9 \\ 
ID & SVO, NA & 47.0 & \bf64.2  & 29.7 & \bf36.9 & 14.6 \\ 
TR & SOV, AN & 60.0 & \bf68.0  & \bf38.8 & 29.3 & 7.9 \\ 
KO & SOV, AN & \bf41.8 & 32.4 & \bf49.3 & 28.8 & 8.0 \\
JA &  SOV, AN & 56.9  &  \bf69.5 & 35.9  &  \bf39.0 &  14.3 \\ \midrule
\multicolumn{2}{c}{Mean SVO} &  50.1 &  \bf71.4 &  33.9 & \bf44.4 & 17.5 \\
\multicolumn{2}{c}{Mean SOV} &  52.8 &  \bf56.7 &  \bf34.1 & 32.4 & 13.9 \\ \midrule
\multicolumn{2}{c}{Mean AN} & 50.6 & \bf66.1 & 34.3 & \bf39.9 & 16.6 \\ 
\multicolumn{2}{c}{Mean NA} &  48.8 &  \bf68.5 & 32.2 & \bf42.6 & 16.3 \\ \bottomrule
\end{tabular}
    \caption{Average dependency accuracy for non-clausal relations (with UD modification) compared with positional baseline. UAS, LAS of constructed trees (w/o UD modification) compared with UAS of left or right branching tree with gold root, whichever is higher. mBERT was used for all languages.}
    \label{tab:ml}
\end{table}
\end{savenotes}

\paragraph{Multilingual Setting}
\phantomsection

\begin{figure}[t]
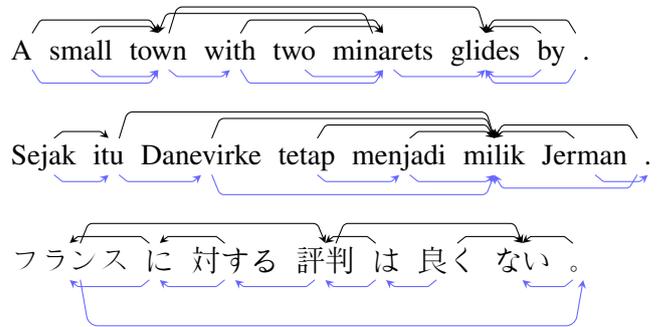

    \begin{flushleft}
    \begin{subfigure}{\linewidth}
        \begin{dependency}[hide label, edge unit distance=.5ex]
		  \begin{deptext}[column sep=0.05cm]
		  A\& small\& town\& with\& two\& minarets\& glides\& by\& . \\
        \end{deptext}
        \depedge{1}{3}{.}
        \depedge{2}{3}{.}
        \depedge{3}{7}{.}
        \depedge{4}{6}{.}
        \depedge{5}{6}{.}
        \depedge{6}{3}{.}
        \depedge{8}{7}{.}
        \depedge{9}{7}{.}
        \depedge[edge style={blue!60!}, edge below]{1}{3}{.}
        \depedge[edge style={blue!60!}, edge below]{2}{3}{.}
        \depedge[edge style={blue!60!}, edge below]{3}{4}{.}
        \depedge[edge style={blue!60!}, edge below]{4}{6}{.}
        \depedge[edge style={blue!60!}, edge below]{5}{6}{.}
        \depedge[edge style={blue!60!}, edge below]{6}{7}{.}
        \depedge[edge style={blue!60!}, edge below]{9}{7}{.}
        \depedge[edge style={blue!60!}, edge below]{8}{7}{.}
        \end{dependency}

    \end{subfigure}
    \begin{subfigure}{\linewidth}
        \begin{dependency}[hide label, edge unit distance=.5ex]
		  \begin{deptext}[column sep=0.05cm]
		  Sejak\& itu\& Danevirke\& tetap\& menjadi\& milik\& Jerman\& . \\
        \end{deptext}
        \depedge{1}{2}{.}
        \depedge{2}{6}{.}
        \depedge{3}{6}{.}
        \depedge{4}{6}{.}
        \depedge{5}{6}{.}
        \depedge{7}{6}{.}
        \depedge{8}{6}{.}
        \depedge[edge style={blue!60!}, edge below]{1}{2}{.}
        \depedge[edge style={blue!60!}, edge below]{2}{3}{.}
        \depedge[edge style={blue!60!}, edge below]{4}{5}{.}
        \depedge[edge style={blue!60!}, edge below]{5}{6}{.}
        \depedge[edge style={blue!60!}, edge below]{3}{6}{.}
        \depedge[edge style={blue!60!}, edge below]{8}{6}{.}
        \depedge[edge style={blue!60!}, edge below]{7}{8}{.}
        \end{dependency}
    \end{subfigure}
    \begin{subfigure}{\linewidth}
    \begin{CJK}{UTF8}{min}
        \begin{dependency}[hide label, edge unit distance=.5ex]
		  \begin{deptext}[column sep=0.05cm]
		  フランス\& に\& 対する\& 評判\& は\& 良く\& ない\& 。 \\
        \end{deptext}
        \depedge{1}{4}{.}
        \depedge{2}{1}{.}
        \depedge{3}{2}{.}
        \depedge{4}{7}{.}
        \depedge{5}{4}{.}
        \depedge{6}{7}{.}
        \depedge{8}{7}{.}
        \depedge[edge style={blue!60!}, edge below]{2}{1}{.}
        \depedge[edge style={blue!60!}, edge below]{3}{2}{.}
        \depedge[edge style={blue!60!}, edge below]{4}{3}{.}
        \depedge[edge style={blue!60!}, edge below]{5}{4}{.}
        \depedge[edge style={blue!60!}, edge below]{6}{5}{.}
        \depedge[edge style={blue!60!}, edge below]{8}{7}{.}
        \depedge[edge style={blue!60!}, edge below]{1}{8}{.}
        \end{dependency}
    \end{CJK}
    \end{subfigure} 
    
\caption{English, Indonesian, and Japanese examples of mBERT extracted trees \textcolor{blue}{edges below}  compared with the correct trees \textbf{edges above}. For Japanese sentence predicted structure is a left branching chain, which is a strong baseline for this language. 
English translation of the sentences: from Indonesian: {\it``The Danevirke has remained in German possession ever since.''}; from Japanese: {\it``France doesn't have a good reputation.''}
} 
\label{fig:extracted-trees}
\end{flushleft}
\end{figure}

In table~\ref{tab:ml} we present the results of our methods applied to mBERT and evaluated on Parallel Universal Dependencies in nine languages. Comparison of the results for English with table~\ref{tab:uas} shows that the dependency accuracy and UAS decreased only slightly by changing the model from enBERT to mBERT, while LAS saw 0.1 pp increase. The model captures syntax comparably well in German, French, and Finnish. 

We observe that results for languages following Subject-Object-Verb (SOV) order (Turkish, Korean, Japanese) are significantly lower than for SVO languages (English, French, Czech, Finnish, Indonesian) in both Dependency Accuracy (14.7 pp) and the UAS (10.5 pp). Our methods outperform the baselines in the latter group by 17.2 pp to 25.4 pp for Dependency Accuracy and from 6.1 pp to 15.5 pp for UAS. The influence of Adjective and Noun order is less apparent. On average, the NA languages results are higher than for the AN languages by 2.4 pp in Dependency Accuracy and 2.7 pp in UAS.

The disparity in the results for SVO and SOV languages was previously observed by \cite{pires2019multilingual}, who fine-tuned mBERT for part of speech tagging and evaluated zero-shot accuracy across typologically diverse languages. We hypothesize that worse performance for SOV languages may be due to their lower presence in mBERT's pre-training corpus.

\section{Multipurpose Heads}

\label{sec:multipurpose}

\begin{figure}[h!]
    \begin{subfigure}{\linewidth}
        \centering
        \includegraphics[width=\linewidth]{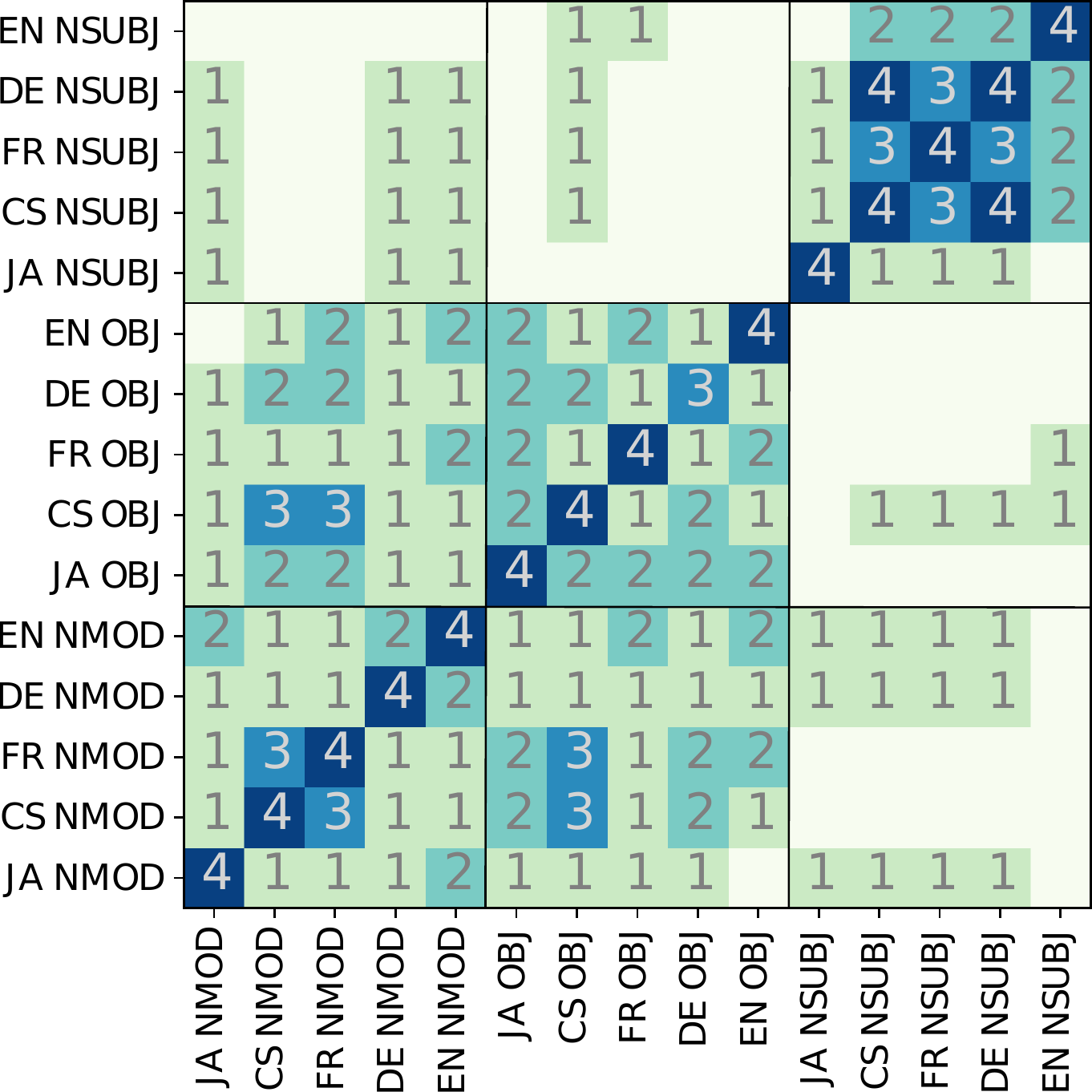}
        \caption{Nominal relations P2D}
        \label{fig:sharedml-nominal}
    \end{subfigure}
    \begin{subfigure}{\linewidth}
        \centering
        \includegraphics[width=\linewidth]{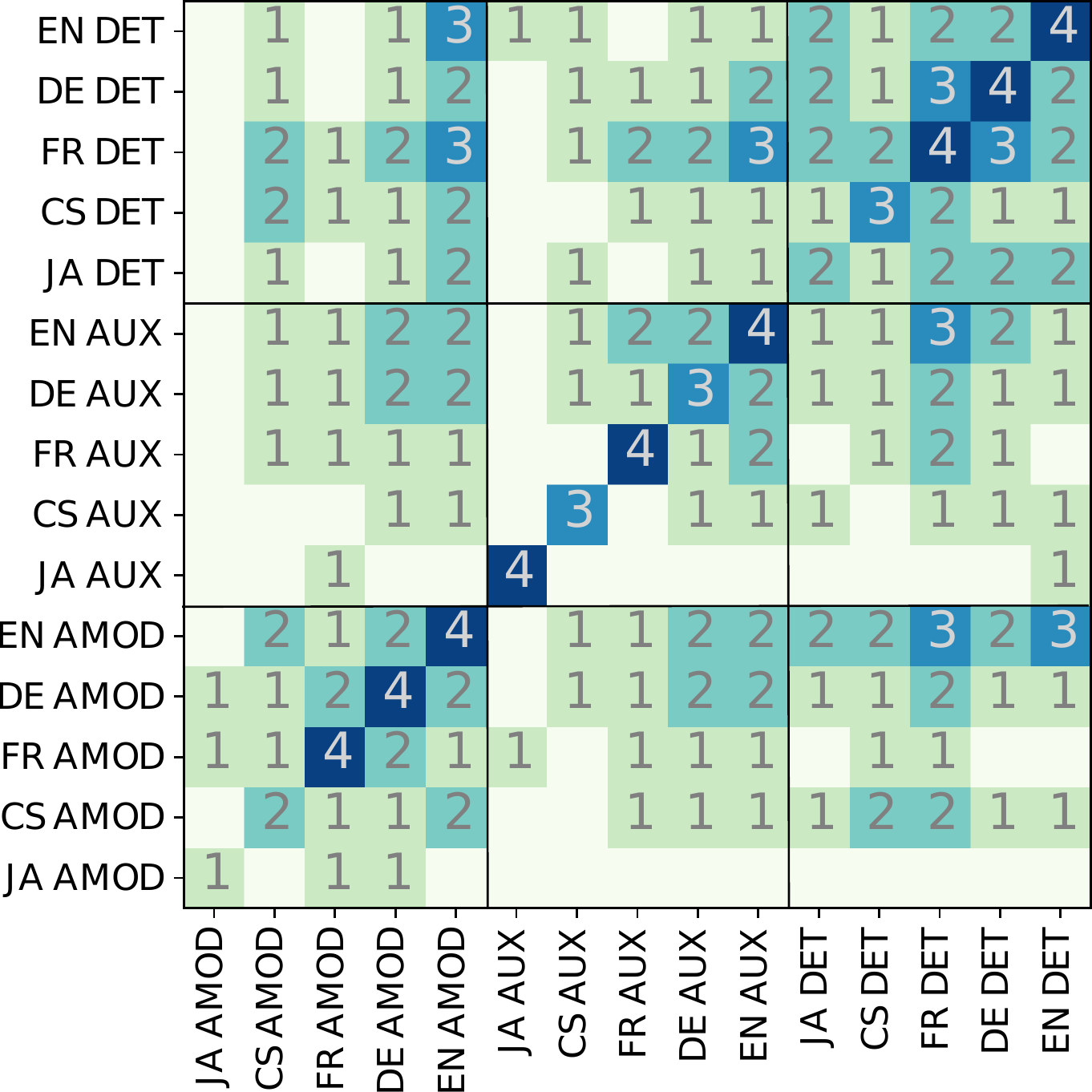}
        \caption{Adjective modifiers, auxiliaries, determiners D2P}
        \label{fig:sharedml-modif}
    \end{subfigure}
\caption{Number of mBERT's heads shared between relations, both within and across languages.} 
\label{fig:sharedml}
\end{figure}

In this experiment, we examine whether a single mBERT's head can perform multiple syntactic functions in a multilingual setting. We choose an ensemble for each syntactic relation for each language. Figure~\ref{fig:sharedml}  presents the sizes of intersections between head sets for different languages and dependency labels.


Except from Japanese, we observe an overlap of the heads pointing to the governor of  adjective modifiers, auxiliaries, and determiners. Shared heads tend to find the root of the syntactic phrase. Interestingly, common heads occur even for relations typically belonging to a verb and noun phrases, such as auxiliaries and adjective modifiers. In our other experiments, we have noticed that these heads do not focus their attention on any particular part of speech.
Similarly, objects and noun modifiers share at least one head for all languages. They have a similar function in a sentence; however, they connect with the verb and noun, respectively. Such behavior was also observed in a monolingual model. 
Figure~\ref{fig:multipurpose} presents attention weights of two heads that belong to the intersection of the adjective modifier, auxiliary, and determiner dependent to parent ensembles.

\subsection{Cross-lingual intersections}
Representation of mBERT is language independent to some extent \cite{pires2019multilingual, libovicky2019language}. Thus, a natural question is whether the same mBERT heads encode the same syntactic relations for different languages. 
In particular, subject relations tend to be encoded by similar heads in different languages, which rarely belong to an ensemble for other dependency labels. Again Japanese is an exception here, possibly due to different Object-Verb order. 

\begin{figure*}[p]

    \centering
    \begin{subfigure}{0.45\linewidth}
    \includegraphics[width=\linewidth]{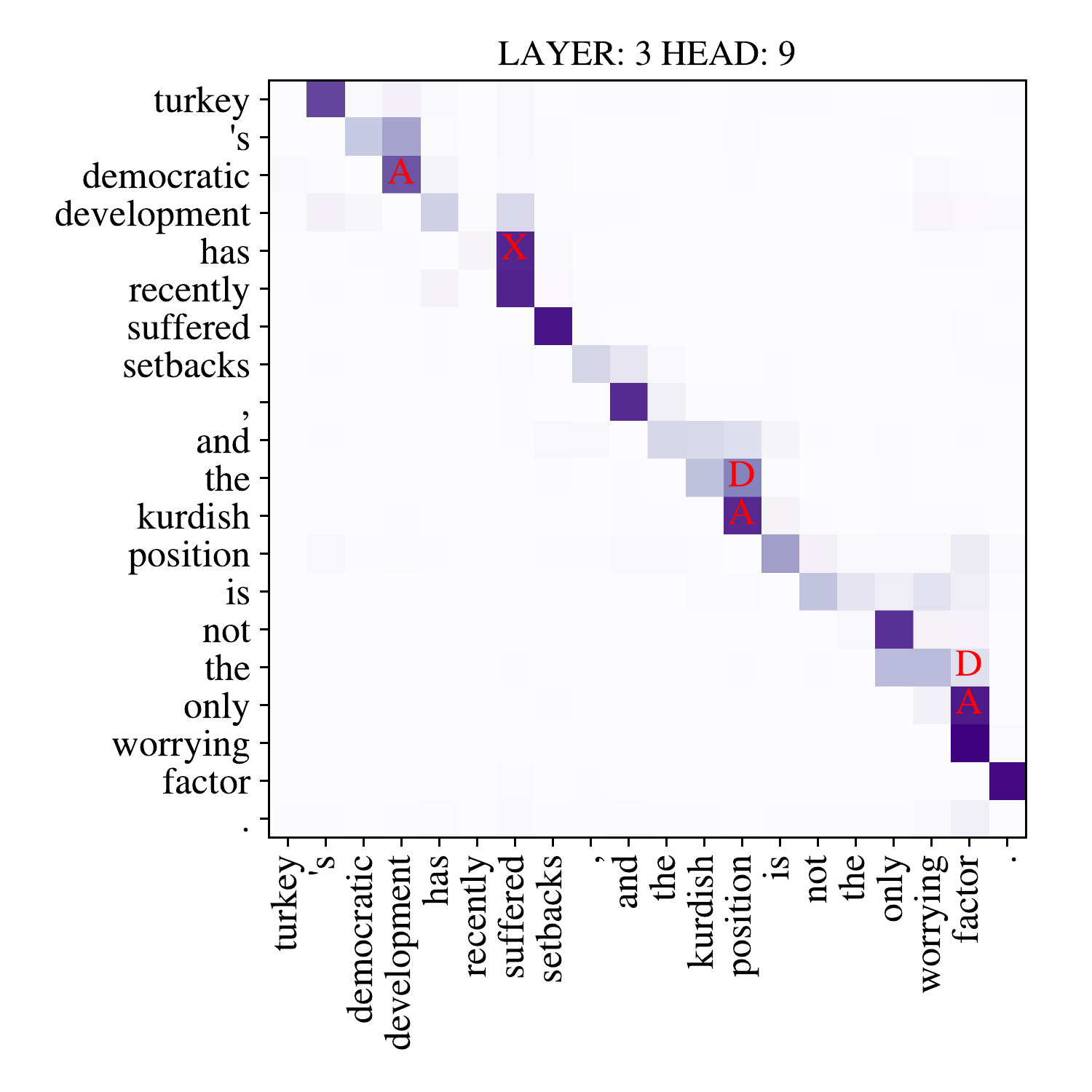}
    \end{subfigure}
    \begin{subfigure}{0.45\linewidth}
    \includegraphics[width=\linewidth]{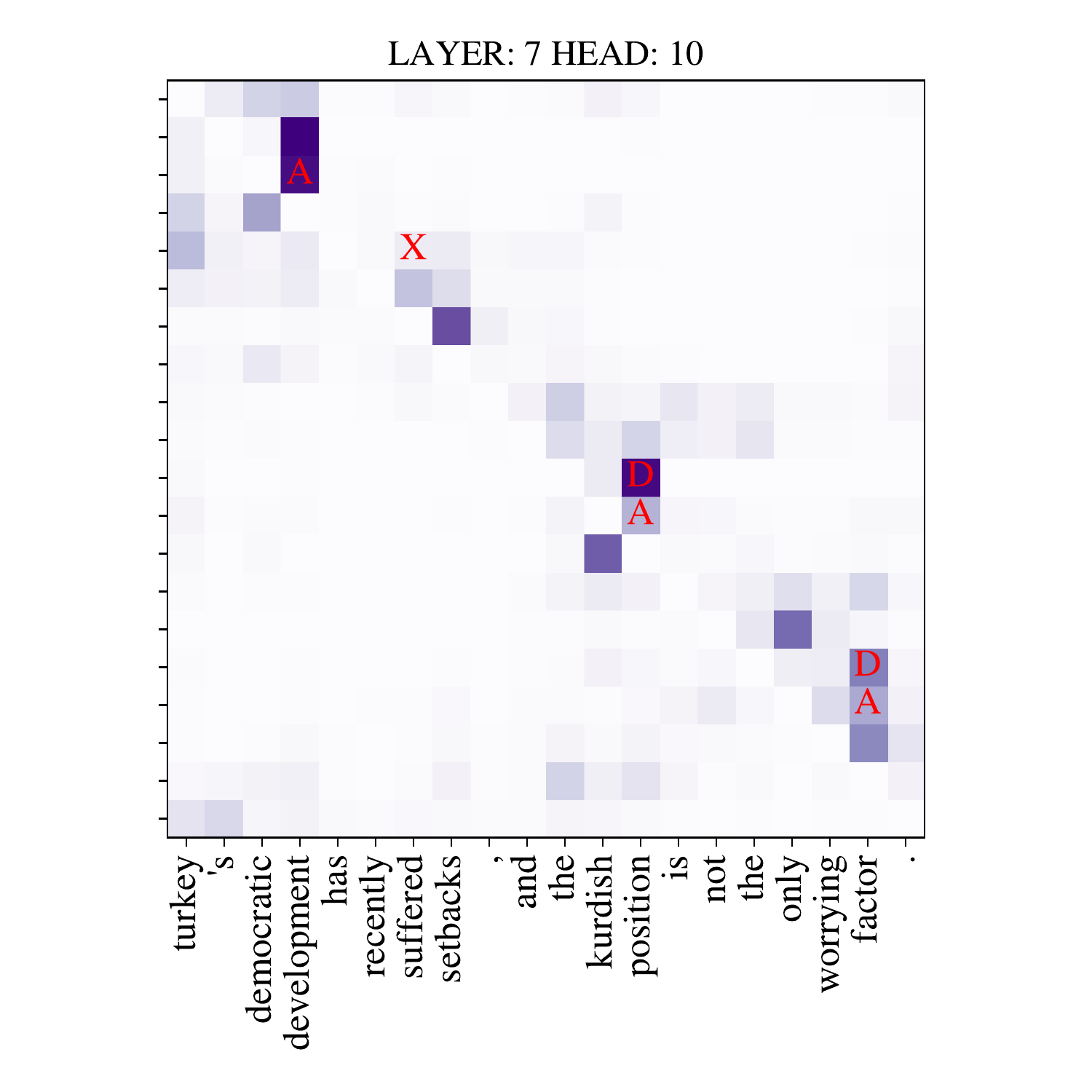}
    \end{subfigure}
    \caption{Syntactic enBERT heads retrieving the parent for three relation labels: \textbf{A}djective modifiers, Au\textbf{X}iliaries, \textbf{D}eterminers. UD relations are marked by A, X, and D respectively.} 
    \label{fig:multipurpose}
    \vspace*{\floatsep}

    \centering
    \begin{subfigure}{0.45\linewidth}
    \includegraphics[width=\linewidth]{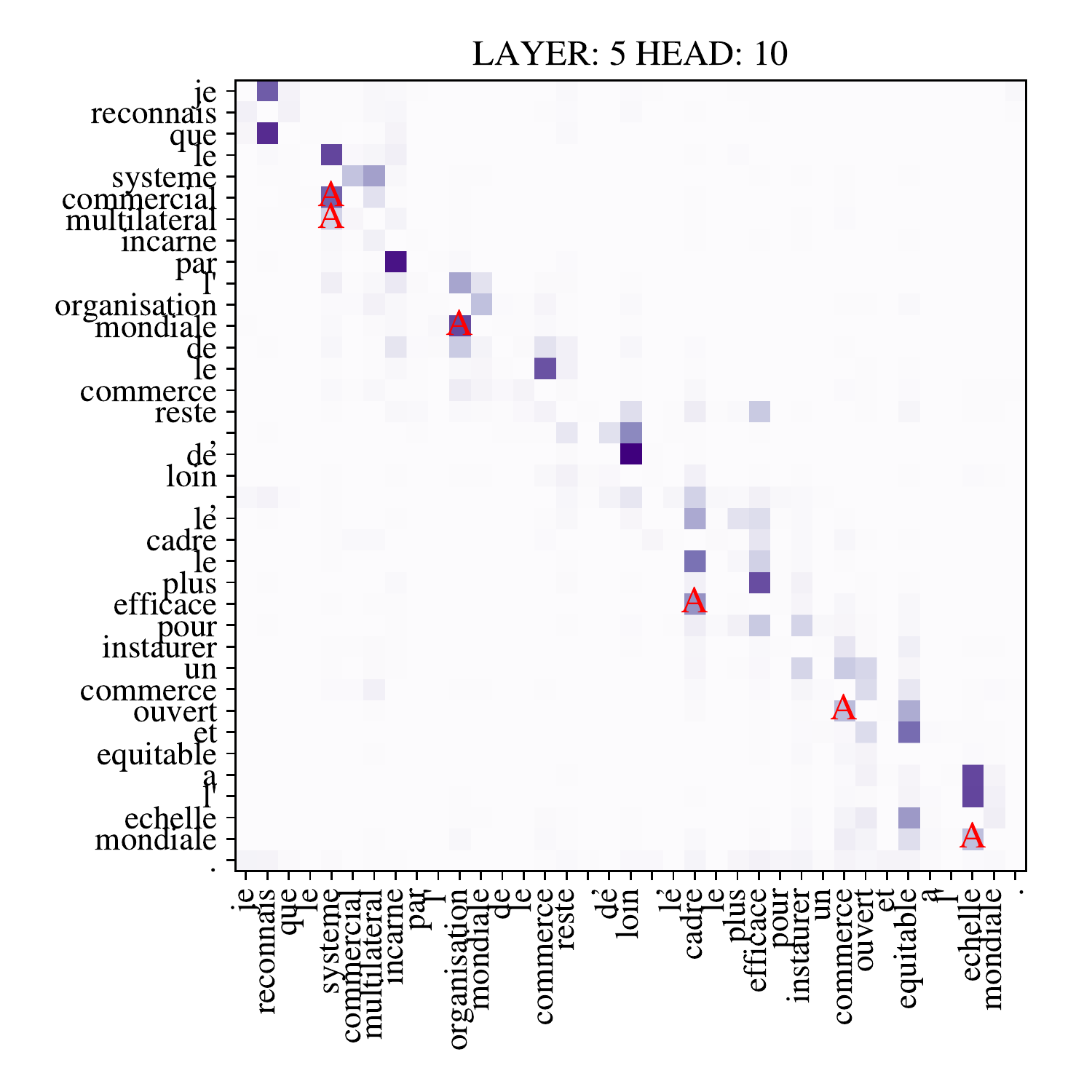}
    \caption{French}
    \end{subfigure}
    \begin{subfigure}{0.45\linewidth}
    \includegraphics[width=\linewidth]{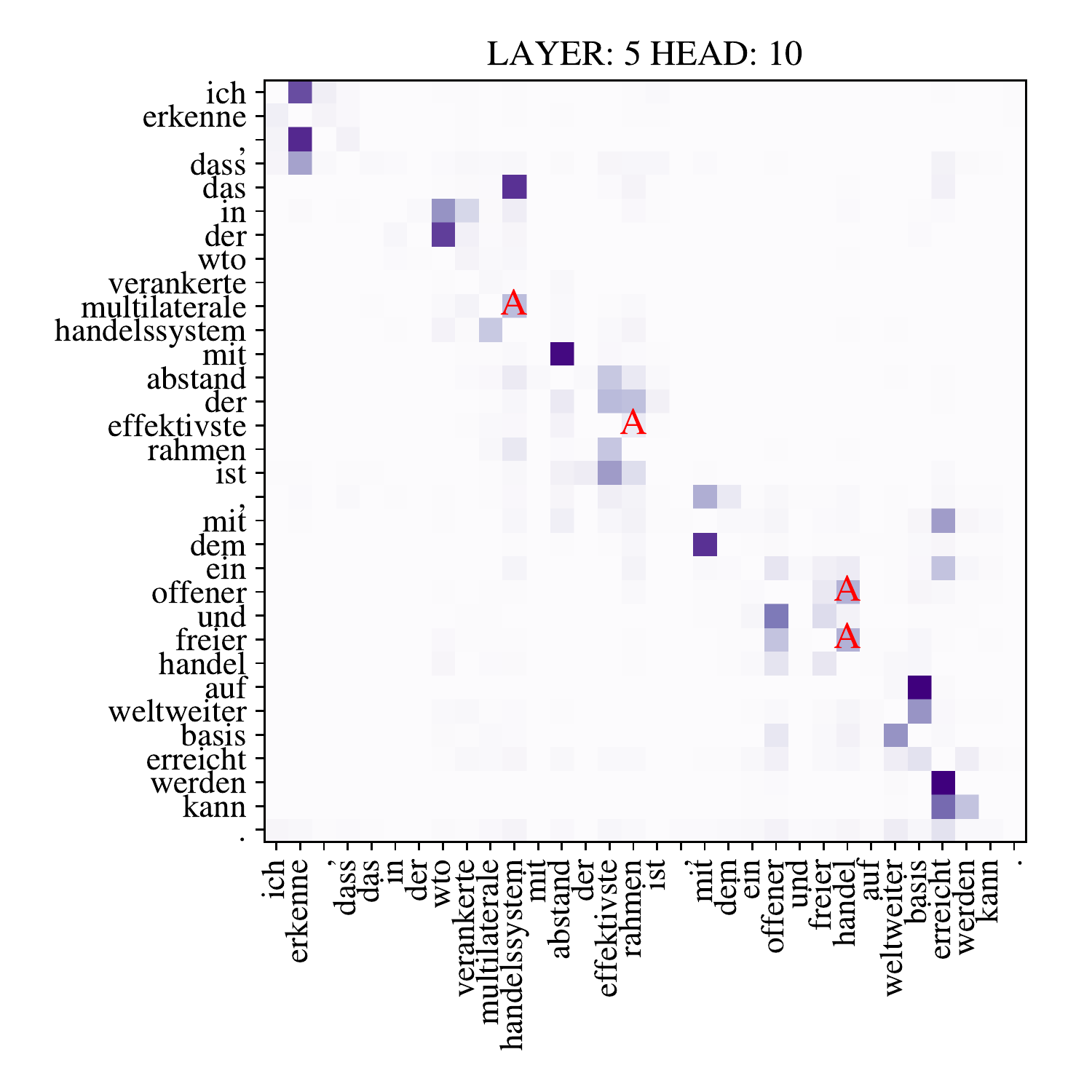}
    \caption{German}
    \end{subfigure}
    \begin{subfigure}{0.45\linewidth}
    \includegraphics[width=\linewidth]{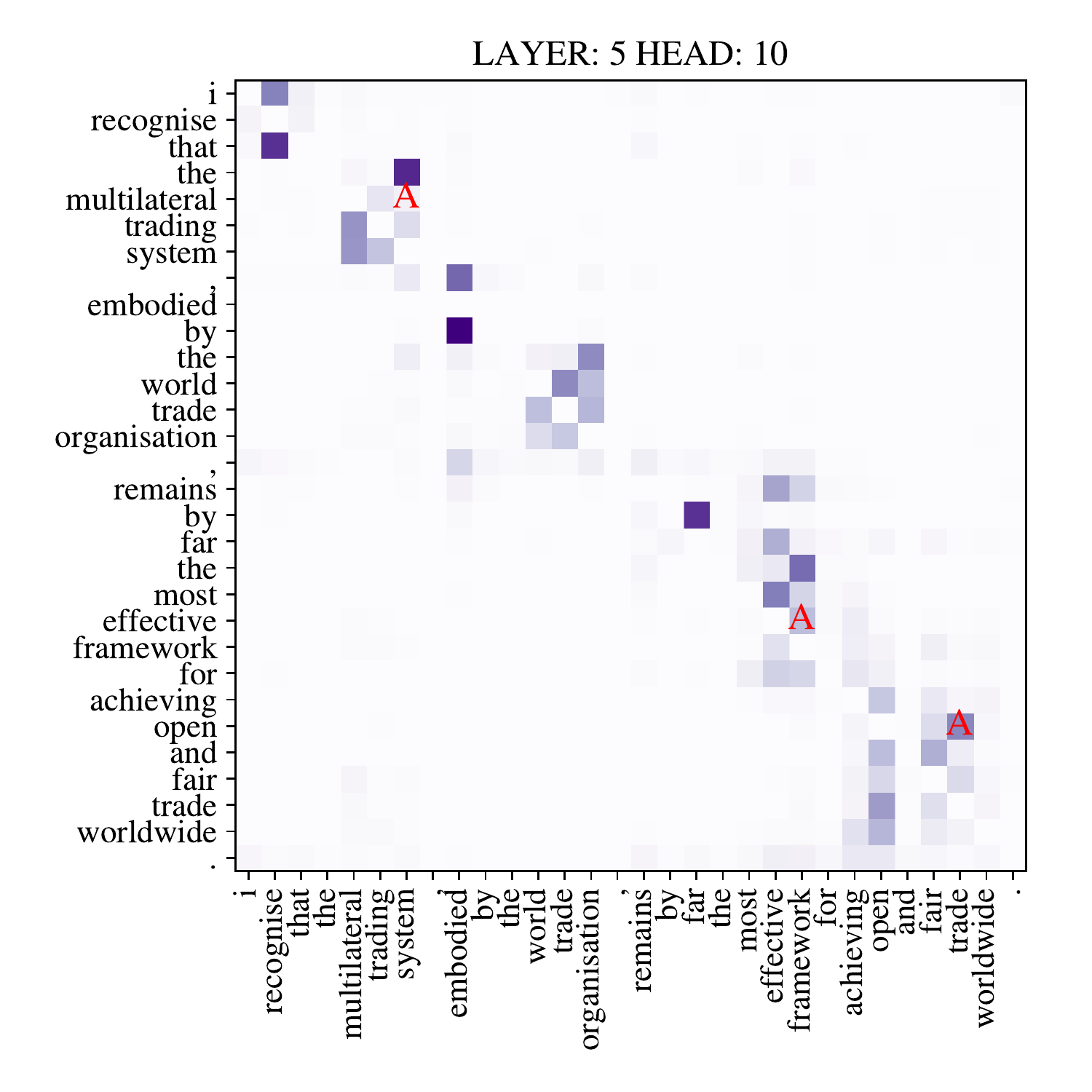}
    \caption{English}
    \end{subfigure}
    \begin{subfigure}{0.45\linewidth}
    \includegraphics[width=\linewidth]{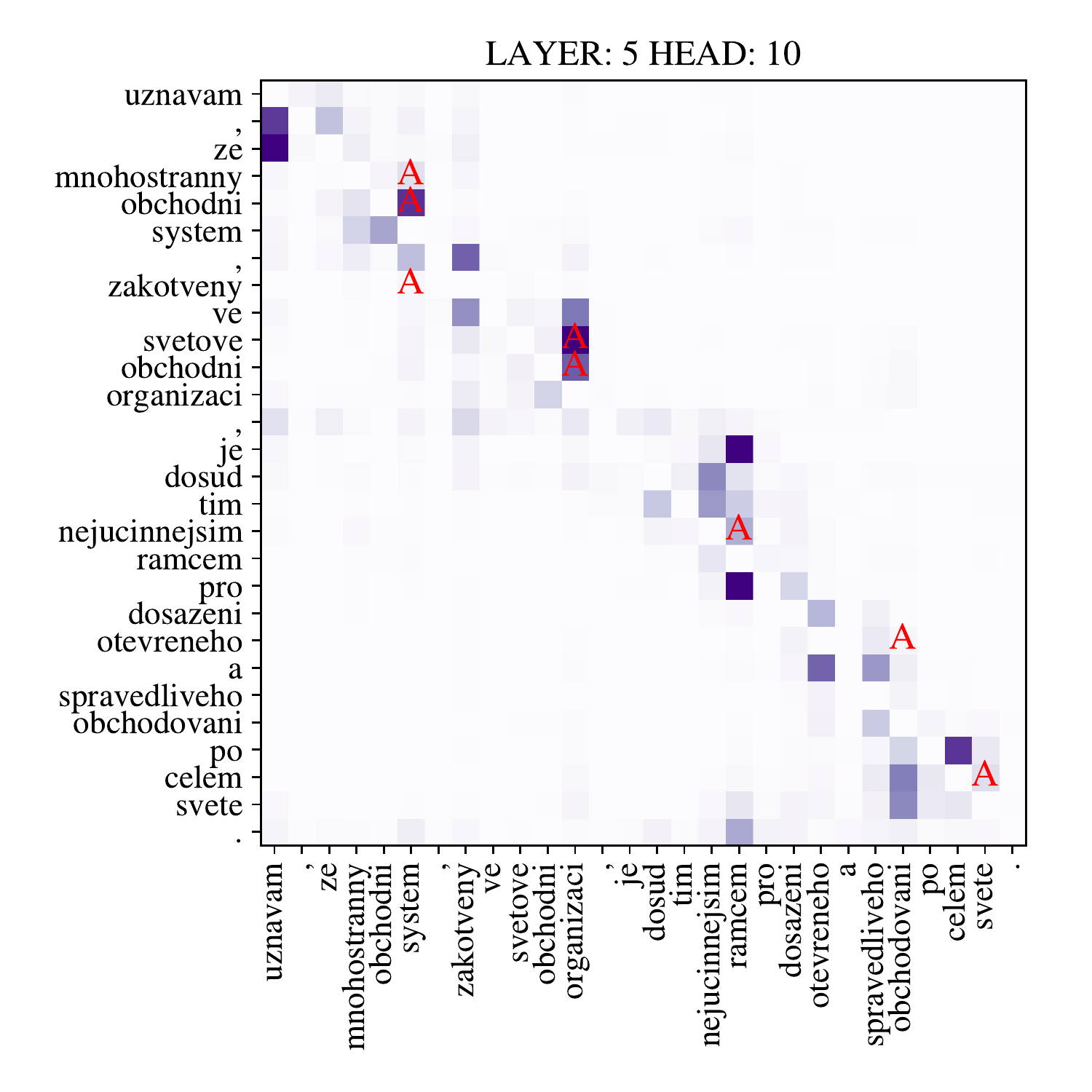}
    \caption{Czech}
    \end{subfigure}
    \caption{A single mBERT head which identifies noun heads of French adjective modifiers. It also partially captures the relation in German, English, and Czech, although these languages, unlike French, follow ``Adjective Noun'' order.}
    \label{fig:multilanguage}
\end{figure*}

For adjective modifiers, the French ensemble has two heads
in common with the German and one with other considered languages,  although the preferred order of adjective and noun is different.
This phenomenon could be explained by the fact that only a few frequent French adjectives precede modified nouns (e.g. ``bon'', ``petit'', ``grand'' ).
Attention weights of a head capturing adjective modifiers in French, German, English, and Czech are presented in Figure~\ref{fig:multilanguage}.

\section{Conclusion}

We have expanded the knowledge about the representation of syntax in self-attention heads of the Transformer architecture.
We modified the UD annotation to fit the BERT syntax better. 
We analyzed the phenomenon of information about one dependency relation type being split among many heads and the opposite situation where one head has multiple syntactic functions.

Our method of head ensembling improved the previous results for dependency relation retrieval and extraction of syntactic trees from self-attention matrices. As far as we know, this is the first work that conducted a similar analysis for languages other than English. We have shown that the method generalizes well across languages, especially those following Subject Verb Object order.

We also hypothesize that the proposed method could improve dependency parsing in a low supervision setting.

\section*{Acknowledgments}

This work has been supported by grant 18-02196S of the Czech Science Foundation.
It has been using language resources and tools developed, stored and distributed by the LINDAT/CLARIAH-CZ project of the Ministry of Education, Youth and Sports of the Czech Republic (project LM2018101)


\bibliography{emnlp2020}
\bibliographystyle{acl_natbib}

\clearpage

\newpage
\appendix

\section{Technical Details}
\subsection{Computing Infrastructure}

We have used one CPU core \textit{Intel(R) Xeon(R) CPU E5-2630 v3} for both head ensemble selection and dependency tree construction. The attention matrices were computed on one GPU core \textit{GeForce GTX 1080 Ti}.

\subsection{Components and Runtimes}

Our pipeline consists of four steps. We provide the average runtime of processing a file of 1000 sentence file for each of them:

\begin{itemize}
    
    \item \textbf{Attention matrices computation} is conducted on GPU for both selection and evaluation sets. 144 self-attention matrices are computed for each sentence and saved in npz file. This step takes approximately 25 minutes.
    
    \item \textbf{Modification of Universal Dependencies} is applied on heads selection and evaluation test (in the latter case only for evaluation of $DepAcc$). We use UDApi with our custom extension (\url{https://udapi.github.io}). The conversion of a CoNLL-U file takes a few seconds.
    
    \item \textbf{Head selection} is done on head selection set. The approximate runtime is 3 minutes.
    
    \item \textbf{Tree extraction} is performed on evaluation set. The approximate runtime is 10 minutes.
    
\end{itemize}

The code is available at GitHub: \url{https://github.com/Tom556/BERTHeadEnsembles}. For details, please refer to the README.
    
\subsection{Data}

Our pipeline requires CoNLL-U files as input. EuroParl parsed sentences used for head selection in English, German, French, Czech, and Finnish are provided in a zip file. 

All other treebanks mentioned in this paper are available at Universal Dependencies webpage \url{https://universaldependencies.org}. 

We perform head selection on the development part of data for Indonesian, Turkish, on train part for Korean and Japanese, due to small amount of development sentences for these two languages.

\section{Original UD Results}

Dependency Accuracy results for English PUD treebank without our modification are presented in the table~\ref{tab:original-ud}.

\begin{table}[!h]
\small
\begin{tabular}{@{}llllll@{}}
\toprule
Relation & Base- & \multicolumn{2}{c}{Orginal} & \multicolumn{2}{c}{Modified}   \\ 
label & line & d2p & p2d & d2p & p2d \\ \midrule
amod & 78.3 & 93.8 & 79.5 & 93.8 & 79.5 \\
advmod & 48.6 & 62.1 & 62.6 & 62.1 & 63.6 \\
aux & 65.2 & 93.4 & 83.1 & 94.5 & 88.0 \\
case & 36.2 & 88.4 & 68.9 & 88.4 & 68.9 \\
compound & 75.8 & 87.0 & 79.1 & 87.0 & 79.1 \\
conjunct & 27.8 & 59.0 & 47.1 & 58.8 & 51.3 \\
det & 56.5 & 97.2 & 69.4 & 97.2 & 69.4 \\
nmod & 25.7 & 49.1 & 54.7 & 49.1 & 54.7 \\
nummod & 57.5 & 79.3 & 72.6 & 79.3 & 72.6 \\
mark & 53.7 & 73.5 & 65.9 & 73.5 & 65.9 \\
obj & 39.2 & 90.8 & 80.7 & 89.3 & 78.5 \\
nsubj & 24.6 & 56.9 & 70.1 & 57.8 & 76.0 \\ \midrule
$\Uparrow$ AVG. \\ NON-CLAUSAL  & 49.1 & \multicolumn{2}{c}{73.4} & \multicolumn{2}{c}{74.1} \\ \midrule
acl & 29.7 & 50.5 & 49.0 & 50.5 & 43.8 \\
advcl & 8.2 & 40.4 & 27.7 & 40.7 & 26.3 \\
csubj & 23.3 & 58.6 & 34.5 & 24.1 & 31.0 \\
x/ccomp & 35.0 & 64.6 & 54.9 & 66.9 & 52.1 \\
parataxis & 4.1 & 16.5 & 13.2 & 23.1 & 24.2 \\ \midrule
$\Uparrow$ AVG. CLAUSAL & 24.7 & \multicolumn{2}{c}{41.0} & \multicolumn{2}{c}{38.3} \\ \midrule
punct & 9.3 & 27.7 & 41.6 & 28.4 & 44.0 \\
dep & 14.2 & 31.7 & 28.1 & 25.1 & 37.0 \\

 \bottomrule
\end{tabular}
\caption{Comparison of dependency accuracy for original and modified UD. Positional baseline was calculated on original UD. The evaluation was done using enBERT's head ensembles of size 4.} 
\label{tab:original-ud}
\end{table}

\begin{figure}[!h]
    \includegraphics[width=\linewidth]{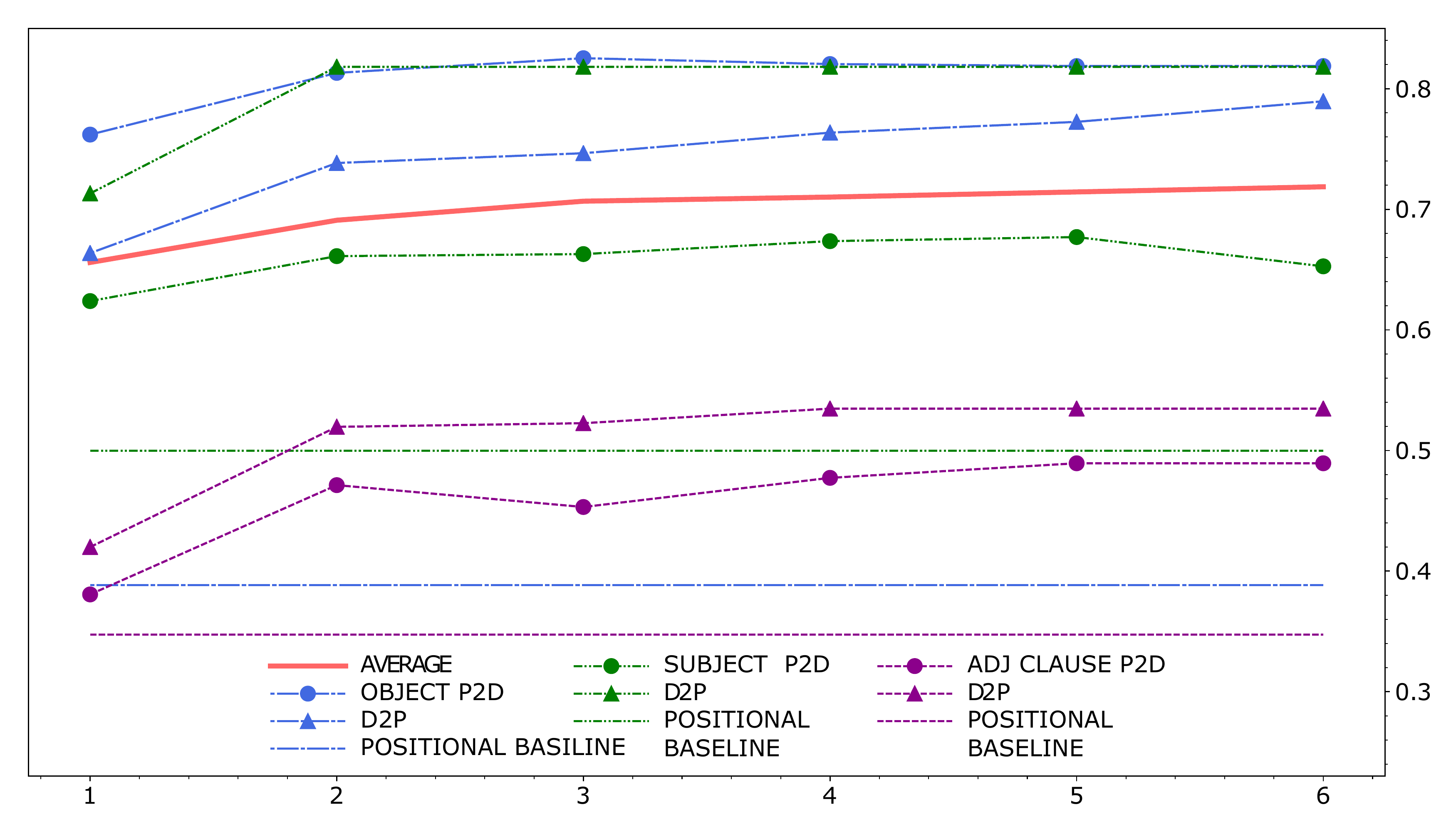}
    \caption{Dependency accuracy on the test set for different sizes of ensembles.} \label{fig:headNum}
\end{figure}

\section{Head Ensemble Size}

In the Figure~\ref{fig:headNum}, we see that ensembles of just two heads have significantly higher dependency accuracy than single heads. For the most relation labels adding more heads does not affect the score, while for a few (object dependent to parent), it grows only slightly. As mentioned in the article, we set the number of heads in ensemble $N$ to 4.

\section{Heads Visualization}

This appendix contains an extended version of the Figure~1 from the article.

\begin{figure*}[]
    \includegraphics[width=\linewidth]{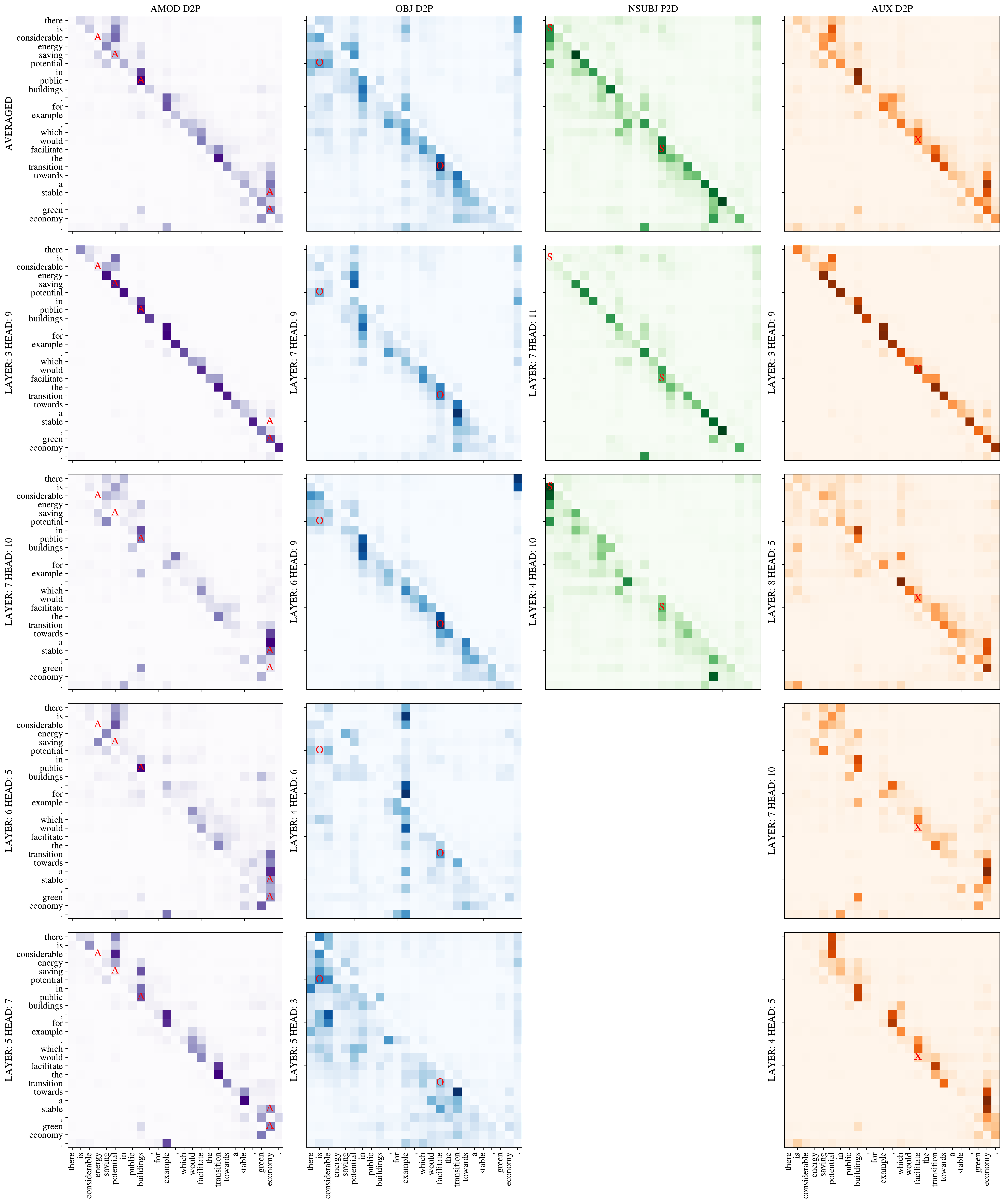}
    \caption{enBERT head ensembles for four dependency  types: adjective modifier (d2p); object (d2p); nominal subject (p2d); auxiliary (d2p). The top row presents averaged attention. UD relations are marked by red crosses. The sentence: ''There is considerable energy saving potential inpublic buildings,  for example,  which would facilitatethe  transition  towards  a  stable,  green  economy.''} \label{fig:attentionsfull}
\end{figure*}



\end{document}